\documentclass[draftcls,svgnames]{IEEEtran}

\usepackage{microtype}
\usepackage{xurl}
\usepackage{listings}
\usepackage{flushend}
\usepackage{marvosym}
\usepackage[hidelinks,pdfauthor={Giuseppe Macario},pdftitle={WebXR, A-Frame and Networked-Aframe as a Basis for an Open Metaverse: A Conceptual Architecture},pdfkeywords={https://orcid.org/0000-0003-4820-155X}]{hyperref}
\usepackage{orcidlink}

\newcommand{\ignore}[1]{[\dots\unkern]} 

\begin{document}
\title{WebXR, A-Frame and Networked-Aframe \\ as a Basis for an Open Metaverse: \\ A Conceptual Architecture}
\author{
\IEEEauthorblockN{\href{https://www.giuseppemacario.men}{Giuseppe Macario}\,\orcidlink{0000-0003-4820-155X}\,\href{mailto:gm@mit.edu.it}{\textcolor{LightSkyBlue}{\Letter}}}

\IEEEauthorblockA{Universitas Mercatorum \\ Piazza Mattei 10, Rome, Italy}
}

\maketitle

\begin{abstract}
This work proposes a WebXR-based cross-platform conceptual architecture, leveraging the A-Frame and Networked-Aframe frameworks, in order to facilitate the development of an open, accessible, and interoperable metaverse. By introducing the concept of spatial web app, this research contributes to the discourse on the metaverse, offering an architecture that democratizes access to virtual environments and extended reality through the web, and aligns with Tim Berners-Lee's original vision of the World Wide Web as an open platform in the digital realm.
\end{abstract}


\section{Introduction}

The advent of extended reality (XR)---namely augmented reality (AR), mixed reality (MR), and virtual reality (AR)---together with a renewed interest in the metaverse after Facebook Inc. changed its name to Meta Platforms Inc., introduces not only unprecedented opportunities for the digital landscape but also significant challenges due to the considerable variation in capabilities between XR devices and conventional computing platforms such as smartphones, tablets, and desktop computers. Where once the primary considerations were screen size and input method (e.g. touch versus mouse and keyboard), developers now must navigate a complex array of sensory inputs, immersive environments and spatial interactions. Therefore, while the web expands to encompass these new immersive experiences, there is a crucial need for the right balance between embracing the unique features of each platform and ensuring universal accessibility. This would allow the web to harness the full potential of XR technologies, enabling creators to deliver rich, immersive experiences while preserving the user's freedom to choose their preferred platform and interface.

Tim Berners-Lee, inventor of the World Wide Web, wanted his creature to be an ``open platform that allows anyone to share information, access opportunities and collaborate across geographical boundaries'' \cite{Solon2017FutureWeb}. To achieve this result, the web should be based on open standards, avoiding proprietary systems \cite{CERN2024HistoryWeb}. The web indeed owes much of its success to the adoption of standards that allow content to be enjoyed by a wide audience, regardless of the browser or the operating system. Due to this inclusive and flexible nature, the web has already become a major platform for consuming two-di\-men\-sion\-al content.

On the other hand, the journey of the web towards a universally accessible platform has not been without its challenges. Instances where proprietary technologies were widely adopted, included but not limited to the ones developed by Microsoft in the 1990s (e.g. ActiveX), exemplify the potential pitfalls of deviating from open standards. This approach led to a significant portion of web content being optimized exclusively for specific browsers, notably Internet Explorer for Windows, thereby limiting accessibility across different platforms. Similarly, the advent of mobile computing devices, including smartphones and tablets, initially posed significant obstacles for web usability due to the vast differences in screen sizes and interaction models compared to traditional desktop computers. However, the web community's commitment to inclusivity and universal access spurred the evolution of standards to overcome these challenges: for example, CSS emerged as a pivotal solution, enabling web content to dynamically adapt to a highly diverse array of devices.

In the growing area of virtual reality and digital interconnectedness, the vision of an open metaverse---a vast, shared, and interoperable virtual space---has captivated the imagination of technologists, creators, and users alike. This work sets out to propose a WebXR-based cross-platform conceptual architecture designed to serve as a cornerstone for building such an open metaverse, delivered through the World Wide Web. Central to this exploration is the research question:

\textit{RQ:} How can a WebXR-based cross-platform conceptual architecture facilitate the development of an open, accessible, and interoperable metaverse, and what are the implications for user engagement and content creation within this digital ecosystem?

The inquiry probes into the viability of leveraging WebXR technology, alongside the integration of the A-Frame and Networked-Aframe frameworks, which enhance the WebXR ecosystem by simplifying the creation of interactive 3D content and enabling real-time, multiplayer experiences across various devices.

\section{Background}

\subsection{Metaverse}
\label{section:metaverse}

When the American writer Neal Stephenson first described a virtual environment named Metaverse in a 1990s sci-fi novel, little did he know that the product of his imagination would soon become reality: Active Worlds (1995) and Second Life (2003) are the first two popular actual forms of Stephenson's Metaverse.\footnote{The Metaverse (with an initial capital letter) is the proper noun of the environment described by Neal Stephenson; in this paper, the metaverse (common noun in lower case) will denote any actual implementation in computer science, distinct from Stephenson's Metaverse.} Since then, the general public has familiarized with the concept of metaverse in information technology, and several computer scientists have given various definitions, for instance
\newtheorem{definition}{Definition}
\begin{definition}[Metaverse]
A (decentralized) three-di\-men\-sion\-al online environment that is persistent and immersive, in which users represented by avatars can participate socially and economically with each other in a creative and collaborative manner, in virtual spaces decoupled from the real physical world \cite{Ritterbusch2023DefiningMetaverse}.
\end{definition}

Updating a prior definition of virtual world \cite{Bell2008VirtualWorlds}, the metaverse can also be defined more concisely as
\begin{definition}[Metaverse]
A persistent synchronous virtual environment, shared by people represented as avatars, facilitated by networked devices.
\end{definition}

These definitions do not explicitly mention XR\slash VR\slash MR\slash AR,\footnote{From now on, XR\slash VR\slash MR\slash AR will be abbreviated to XR.} and do not require interoperability between different metaverses, which makes sense because most implementations, especially the older ones (including the aforementioned Active Worlds and Second Life), do not have XR capabilities and are incompatible with each other. However, after 30 years since the inception of the metaverse, technology is mature enough to handle these new demands \cite{EU2023NextGenVirtualWorlds}. This is where WebXR, A-Frame and Networked-Aframe come in.

\subsection{WebXR}
\label{section:webxr}

The W3C Immersive Web Working Group has been established to address the challenges of crafting a standard named WebXR Device API; its stated mission is ``to help bring high-performance virtual reality (VR) and augmented reality (AR), collectively known as extended reality (XR), via APIs to interact with XR devices and sensors \emph{in browsers}'' \cite{W3C2024ImmersiveWebCharter}.  As a matter of fact, the centrality of browsers as gateways to digital information has remained a constant since the inception of the World Wide Web: recognizing this pivotal role, the W3C has made a strategic decision to leverage the browser as a foundational platform for spatial web apps that render immersive environments. WebXR spatial web apps can run on any device equipped with a modern browser, enabling a wide audience to engage with 3D environments without necessarily resorting to specialized software or hardware.

The WebXR Device API is designed to offer a unified, platform-agnostic abstraction layer that grants access through the web to the fundamental features shared by all XR devices. By abstracting the complexities of underlying hardware, not only does the WebXR Device API enable web developers to access a broad spectrum of interaction controllers through a streamlined interface, but it also simplifies the real-time rendering of the immersive environments.

WebGL, a JavaScript API that leverages the capabilities of OpenGL for Embedded Systems (OpenGL ES) to render interactive 2D and 3D graphics directly within web browsers, serves as the foundational layer for graphical rendering in WebXR applications. Through its direct access to the GPU, WebGL provides the low-level interface to the GPU, enabling high-performance rendering of complex 3D graphics and animations. In order to abstract away the complexities of direct WebGL programming, WebXR applications typically use a higher-level open-source framework or library, to choose from A-Frame, Three.js, or Babylon.js.

\begin{figure}
\centering
\includegraphics[width=\columnwidth,alt={360° photorealistic view of a bedroom}]{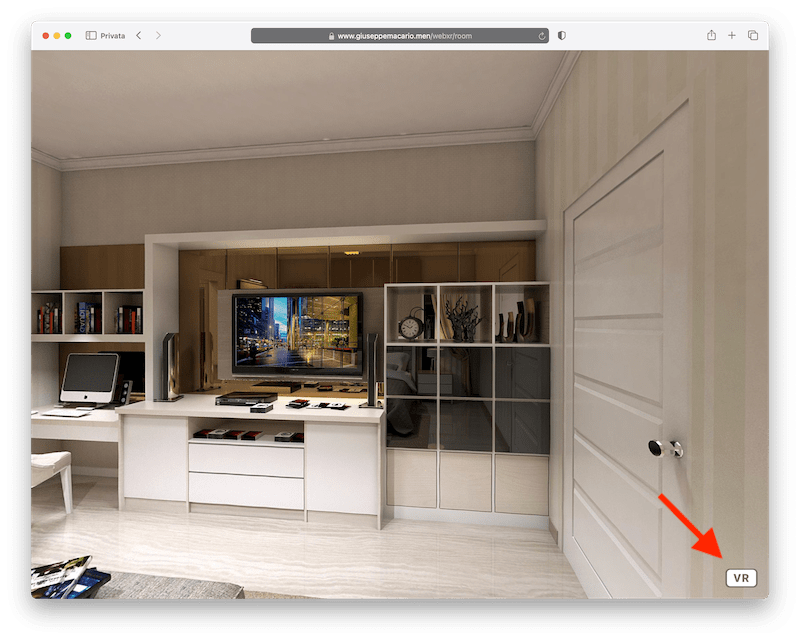}
\caption{VR button in the lower right corner of an A-Frame scene}
\label{fig:room}
\end{figure}

A constantly updated list of browsers supporting WebXR is available online\footnote{\url{https://caniuse.com/webxr}} and, at the time of writing (\the\year{}), it shows that the latest versions of Chrome, Edge, Opera, and Samsung Internet, fully support the API. Firefox and Safari also support the API---not yet on iOS\footnote{In the meantime, WebXR is supported by other iOS apps such as Mozilla's WebXR Viewer.}---although it may have to be manually enabled by the user. Consequently, all the most popular XR headsets also support WebXR, through their built-in browsers: Meta Quest, Microsoft HoloLens, Apple Vision Pro, HTC Vive, Samsung Gear, Google Cardboard and others. Interestingly, desktop and mobile browsers also support WebXR; however, since desktops, laptops, smartphones and smart TVs are not specifically designed for XR, the browser is forced to graphically simulate the immersive features of WebXR, depending on the screen and the platform. This is why A-Frame\footnote{See section \ref{section:a-frame}.} apps can show a button in the lower right corner (fig.~\ref{fig:room})
that enables users to transition from a traditional 2D web environment into a fully immersive 3D experience. This critical interface element seamlessly integrates the digital content of the application with the immersive capabilities of the XR device, providing users with a more engaging and interactive experience. Depending on the needs of the developer, the button can be displayed on any browser supporting WebXR; in the case of desktop browsers, it toggles full screen mode.

\subsection{A-Frame}
\label{section:a-frame}

By abstracting the Three.js JavaScript library into an HTML-like syntax through the entity-component-system (ECS) architectural pattern \cite{Wang2020CrossPlatformWebBasedVR}, the A-Frame framework enables easy programming of WebGL applications. The advantages of the ECS pattern, which is popular in 3D and game development, and the reasons why it has been adopted by A-Frame, are described in \cite{Super2024ECS}. The embedding of the WebXR application programming interface (API) makes A-Frame compatible with XR devices. Through a web browser, the resulting web applications can be used with common XR systems as well as conventional PC and mobile devices. In this way, the largest possible number of users is reached without requiring the use of special hardware or the installation of software. With A-Frame, various components can be used to create interactive virtual worlds with virtual objects and 3D models, lighting, material, and multimedia assets such as images, videos, and sounds. For expanding the functionality of the framework, custom A-Frame components can be programmed in JavaScript.

\subsection{Networked-Aframe}

Since A-Frame is primarily geared towards single-user experiences, the Networked-Aframe framework builds upon A-Frame by introducing networked components that enable real-time, multi-user experiences within the same virtual space, mirroring the dynamics of physical users' interactions. Networked-Aframe communicates via WebRTC, a peer-to-peer (P2P) API, or via Web\-Sockets, a client-server API, facilitating low-latency communication, which is critical for maintaining the illusion of presence and ensuring a responsive user experience.

\section{Previous work}

Researchers started to identify WebXR, A-Frame, and Networked-Aframe as promising technologies for the metaverse in 2023, albeit with limited exploration in scholarly articles so far. Earlier works about Networked-Aframe and virtual worlds such as \ignore{Mozilla Hubs} (without mentioning the metaverse) date back to 2019.

Dziwis, von Coler, and Porschmann \cite{Dziwis2023LiveCoding} ventured into the realm of networked, immersive, and shared virtual environments for live coding performances, introducing two browser-based live coding languages for real-time, collaborative experiences within metaverse systems. This initiative not only showcases the utility of metaverse environments for artistic expression but also highlights the growing interest in integrating more interactive and immersive elements into these platforms. At the same time, the authors developed Orchestra \cite{Dziwis2023Orchestra}, an open-source toolbox designed for live music performances within web-based metaverse environments, which were further explored by Tomasetti et al. \cite{Tomasetti2023SpatialAudio}.

Sobota et al. \cite{Sobota2023VirtualEducational} presented a web-based educational cooperative virtual environment tailored for low-cost mobile VR headsets. This approach addresses the accessibility and cost barriers in educational VR, suggesting a wider applicability of these technologies in resource-constrained settings.

Earlier works by Scavarelli et al. \cite{Scavarelli2019Framework} already recognized the importance of inclusive and social VR in educational settings, proposing frameworks that support accessible and collaborative content within social learning spaces. Their efforts culminated in Scavarelli's thesis on a multi-platform virtual reality framework aimed at overcoming VR's inclusion problem in learning environments \cite{Scavarelli2023InclusiveVR}.

Collectively, these studies underscore a growing trend towards the diversification of metaverse applications, ranging from music and art to education and collaborative workspaces. Despite the nascent state of research in this area, the body of work demonstrates a clear trajectory towards more immersive, interactive, and accessible metaverse experiences. The integration of advanced user interaction components, the development of tools for artistic expression, and the focus on educational and accessible VR solutions reflect a comprehensive effort to leverage WebXR, A-Frame, and Networked-Aframe technologies for a broad spectrum of metaverse applications.

\section{Spatial web apps}

The academic introduction of the term spatial computing is attributed to \cite{Greenwold2003SpatialComputing}:
\begin{definition}[Spatial computing]
Human interaction with a machine in which the machine retains and manipulates referents to real objects and spaces.
\end{definition}

In 2023, Apple stated: ``Featuring visionOS, the world's first spatial operating system, Vision Pro lets users interact with digital content in a way that feels like it is physically present in their space'' \cite{Apple2024SpatialComputing}. Therefore, a spatial app is an app that needs a spatial operating system such as visionOS.

This paper introduces the concept of spatial web applications, i.e. spatial apps that do not need to run on a spatial operating system. To run a spatial web app, users merely have to open a URL. This can pave the way for creative frictionless experiences: for example, a user can scan a QR code (representing a URL) on a real world poster to launch a spatial WebXR app that transforms the poster into an XR app. Although no installation is required, app installation is still possible via features such as Progressive Web Apps (PWA) and ``Add to Home Screen'', which allows mobile websites to open like native avoiding the Apple or Google App Stores. An app delivered as a regular web page will never have to be manually updated by the user, because it is sufficient for the developer to update the web page. Moreover, the app automatically benefits from browser distinctive features such as history or tabs, which are already well known to the general public.

\begin{figure}
\centering
\includegraphics[width=\columnwidth,alt={Floating virtual terrestrial globe in a real room}]{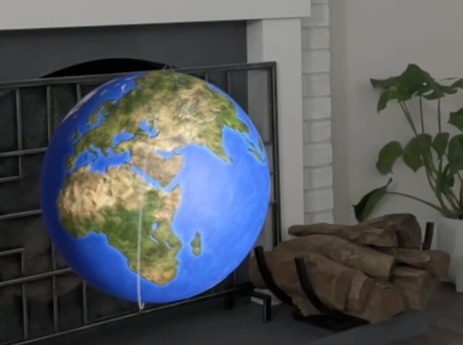}
\caption{``Hello world'' in visionOS and WebXR}
\label{fig:hello-world}
\end{figure}

Apple provides an example of ``Hello world'' in visionOS \cite{Apple2024HelloWorld}, i.e. a spatial app that shows a floating globe (Fig.~\ref{fig:hello-world}); this downloadable Xcode project contains 180 files, for a total of about 250~MB of source code.
However, Apple Vision Pro also supports WebXR through Safari for visionOS, which allows the developer to create a spatial app without using Xcode and avoiding Apple's review process. Instead of compiling a project of 180 files, WebXR does the same thing in very few lines of A-Frame code---even just one line---within one HTML document:
\begin{lstlisting}[language=HTML, basicstyle=\footnotesize\ttfamily, showstringspaces=false, breaklines, flexiblecolumns, keywordstyle={}]
<body>
  <a-scene>
    <a-sphere position="0 1.5 -5" radius="1"
      rotation="0 0 -30"
      src="url(https://example.com/texture.jpg)"
      animation="property: rotation; to: 0 360 -30; loop: true; dur: 10000; easing: linear">
    </a-sphere>
  </a-scene>
</body>
\end{lstlisting}

In general, a spatial web app basically consists of at least an HTML document with JavaScript, textures and 3D models. A minimal valid HTML page including the aforementioned code is available online,\footnote{\url{https://www.giuseppemacario.men/webxr/hello-world}} where devices can also test the app.

All in all, this example shows some of the advantages of a WebXR spatial web app over a native visionOS app. Firstly, the app can run on any device and not just on Apple Vision Pro. Secondly, the source code of an HTML document is extremely light compared to a huge Xcode project, and does not need to be compiled. Furthermore, the spatial web app does not have to be distributed through an app store. The same advantages hold true for other non-Apple headsets.

\section{Conceptual architecture}

\begin{figure}
\centering
\includegraphics[width=\columnwidth,alt={Stack of layers from the bottom up: JavaScript, WebGL (OpenGL ES), WebXR Device API, Three.js, A-Frame, Networked-Aframe}]{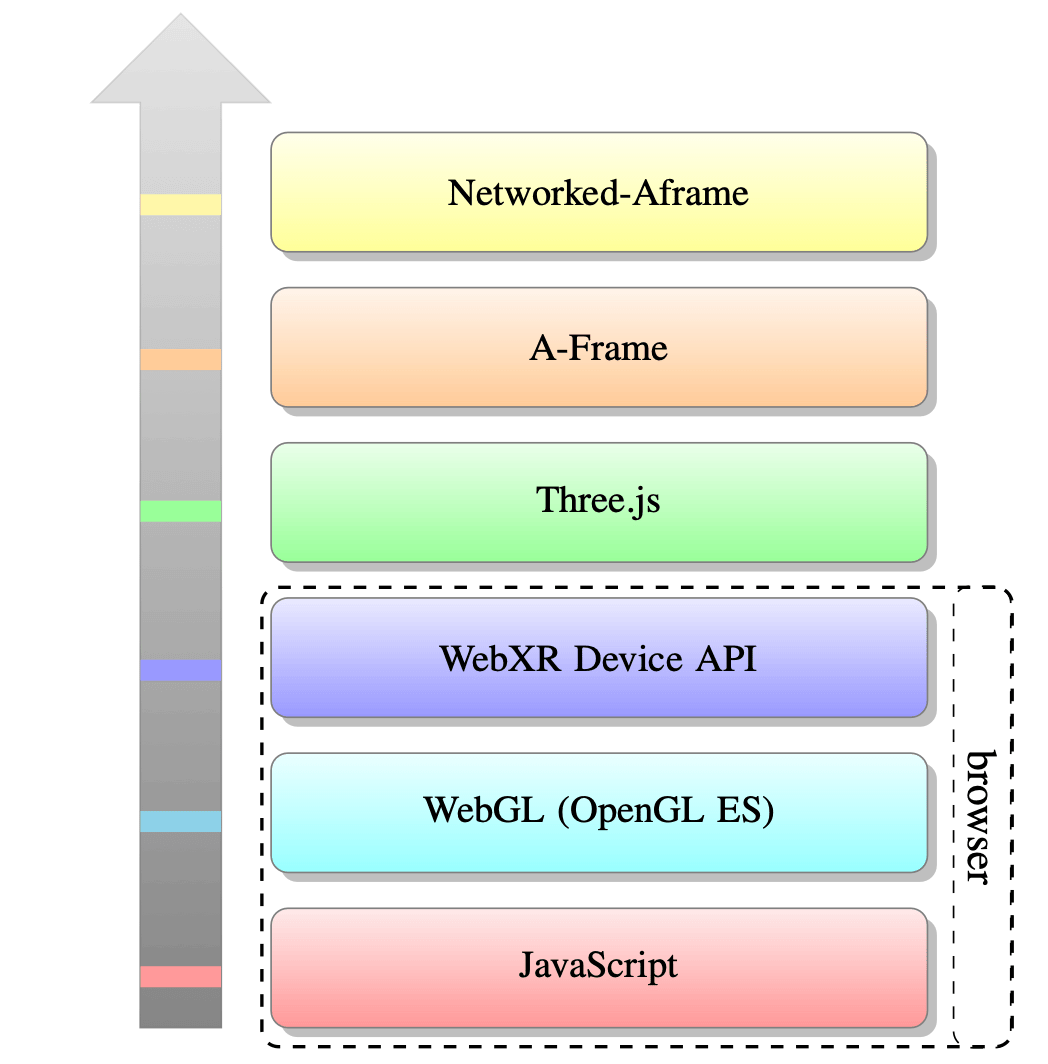}
\caption{Technology stack of a WebXR\slash Networked-Aframe spatial web app (abstraction layers in ascending order)}
\label{fig:webxr-stack}
\end{figure}

The conceptual architecture, as illustrated in fig.~\ref{fig:webxr-stack}, represents a layered approach to developing spatial web applications. Each layer builds upon the foundational technologies: \\
\textit{1. JavaScript:} Serves as the primary programming language, enabling the development of interactive web applications. \\
\textit{2. WebGL (OpenGL ES):} Provides a low-level graphics API for rendering 2D and 3D graphics in the browser. \\
\textit{3. WebXR Device API:} Facilitates direct interaction with XR hardware. \\
\textit{4. Three.js:} Acts as a lightweight 3D library that builds on top of WebGL, simplifying 3D content creation. \\
\textit{5. A-Frame:} Provides a web framework for building VR experiences, making Three.js accessible to web developers without deep knowledge of WebGL. \\
\textit{6. Networked-Aframe:} Extends A-Frame for multiplayer experiences, enabling the development of a shared metaverse.

As seen in section \ref{section:webxr}, the first three components are natively supported by web browsers; therefore, in fig.~\ref{fig:webxr-stack}, they are encapsulated within a ``browser'' environment, indicating the stack's capability to deliver immersive experiences directly through a web browser without the need for external applications.

\section{Features of a WebXR-based metaverse}

According to the definitions provided in section~\ref{section:metaverse}, persistence and multi-user concurrency are two key features of the metaverse. Besides, \ignore{Mozilla}'s engineers described several ``open problems'' of WebXR \ignore{\cite{MacIntyre2018FutureWebXR}}, which can be reformulated and classified into five groups: immersive web browsers, computer vision, geolocation, persistence, and multi-user concurrency. Let us see how these critical issues can be addressed and overcome in a state-of-the-art WebXR-based metaverse.

\subsection{Immersive browsers}

In 2022, after WebXR was integrated into all the most popular web browsers, Mozilla ceased to maintain a browser specifically developed for VR devices, named Firefox Reality: an additional immersive browser is no longer considered essential because all the major browsers, on all the major operating systems,\footnote{Except for Safari on iOS, as seen in section~\ref{section:webxr}.} are now able to provide immersive experiences, as indicated in the bottom layers of fig.~\ref{fig:webxr-stack}.

\subsection{Persistence}

In the metaverse, persistence refers to the ability to maintain the state, properties and evolution of digital assets, as well as user progress and interactions across sessions and platforms. Therefore, persistence necessitates a multifaceted approach leveraging different storage modalities.

\paragraph{Centralized storage} This storage modality operates on the back end, typically hosted in the cloud or on dedicated servers, facilitating centralized management of data that needs to be persistently available across sessions and platforms. Server-side databases, whether SQL-based (such as PostgreSQL or MySQL) for structured data or NoSQL databases (such as MongoDB, Couchbase, or DynamoDB) for unstructured or semi-structured data, offer scalable solutions for managing complex datasets inherent in virtual worlds.

\paragraph{Client-side storage} Developers can implement transient state persistence directly on the client's device by using JavaScript APIs provided by browsers, namely IndexedDB and Web storage (or DOM storage), which have both been standardized by the W3C. This approach facilitates rapid access to data with minimized latency, crucial for session-specific data like user preferences and session progress.

\paragraph{Distributed storage} An open metaverse can benefit from a free and global solution for storing and accessing assets such as 3D models, textures, and user-generated content. The InterPlanetary File System (IPFS) is a great candidate: unlike centralized cloud storage, it operates in a fully distributed manner. However, without incentives or arrangements for pinning services, there is no guarantee that data will remain available indefinitely on IPFS, because data is only retained as long as at least one node in the network keeps hosting it.

\subsection{Multi-user concurrency}

A-Frame provides a robust platform for the development of VR environments using HTML-like syntax, simplifying the creation of 3D and VR content on the web. However, its core functionality is primarily geared towards single-user experiences. The Networked-Aframe framework builds upon the A-Frame framework by introducing networked components that enable real-time, multi-user experiences within the same virtual space, mirroring the dynamics of physical interactions. Networked-Aframe communicates via WebRTC, a peer-to-peer (P2P) API, or via WebSockets, a client-server API. Networked-Aframe facilitates low-latency communication, which is critical for maintaining the illusion of presence and ensuring a responsive user experience.

Networked-Aframe incorporates several key features that are essential for implementing multi-user concurrency in the metaverse:

\paragraph{Real-time synchronization} Networked-Aframe synchronizes the state of objects across all connected clients in real-time. This ensures that any interaction or modification of the virtual environment by one user is immediately reflected to all other users, providing a cohesive and interactive experience.

\paragraph{User presence and interaction} Networked-Aframe supports the representation of users within the virtual space, often through avatars, allowing for the visualization of other users' presence and movements. This feature is crucial for social interactions and collaborative activities within the metaverse.

\paragraph{Networked communication} Beyond visual synchronization, Networked-Aframe facilitates networked communication channels, enabling users to share messages, voice, and even video within the virtual environment. This enhances the depth of interaction, making the virtual space more engaging and immersive.

\paragraph{Scalability} The framework is designed with scalability in mind, allowing for the efficient management of network traffic and state synchronization among a large number of users. This scalability is vital for the expansive environments envisioned in the metaverse.

\subsection{Computer vision}

Although WebXR does not inherently include computer vision capabilities as part of its core API, it can work in conjunction with several open-source AR-oriented libraries that offer image tracking, face tracking and marker tracking. When necessary, these libraries rely on OpenCV.js and Tensorflow.js, which provide robust vision processing and machine learning capabilities.

\paragraph{OpenCV.js} This is a JavaScript binding for a selected subset of OpenCV's comprehensive functions tailored for web platforms: it enables developers to harness a wide array of sophisticated vision processing capabilities directly within WebXR applications, ranging from image manipulation and object detection to more complex operations like facial recognition and markers detection. Not only does the integration of OpenCV.js with WebXR simplify the development process by providing a unified web-centric API for complex vision tasks, but it also significantly enhances the interactivity and realism of WebXR experiences, allowing for dynamic and responsive environments that react to the user's physical environment and inputs.

\paragraph{TensorFlow.js} As the JavaScript implementation of a renowned open-source library for machine learning and artificial intelligence, TensorFlow.js provides a comprehensive suite of tools and APIs for training and deploying machine learning models within a WebXR spatial app through the browser. This integration facilitates the creation of highly interactive and intelligent WebXR experiences, where machine learning algorithms can process and interpret real-time data from the user's environment, such as visual cues, gestures, and voice commands. By leveraging TensorFlow.js, WebXR applications can achieve advanced functionalities like gesture recognition, object detection, and environmental understanding, without the need for external servers or specialized hardware. TensorFlow.js thus represents a key technology in expanding the capabilities of WebXR, making it possible for developers to create sophisticated adaptive AI-enhanced immersive experiences directly in the browser.

The landscape of WebXR development has therefore transcended the barrier of integrating complex computer vision capabilities. The challenges of incorporating computer vision into WebXR are no longer insurmountable technical hurdles but rather exciting opportunities to innovate and enhance the way the user interact with AR and VR. Through the continued advancement and integration of these libraries, WebXR stands at the forefront of a revolution in immersive web technology, making the future of virtual and augmented reality on the web not just accessible but boundlessly creative and interactive.

\subsection{Geolocation}

While the core WebXR Device API does not provide direct geolocation capabilities as of its current specification, it allows for the integration of geolocation data through the use of additional APIs and sensor data accessible in the web environment:

\paragraph{Geolocation API} This web API provides access to geographical location information from the browser. It can be used alongside WebXR to create location-aware AR experiences. By fetching the user's latitude, longitude, and altitude, developers can integrate real-world positions with virtual environments.

\paragraph{AR.js} Focused specifically on bringing AR capabilities to the web, AR.js works well with WebXR and provides features for marker-based and location-based AR. It simplifies the process of integrating digital content with real-world coordinates, making it easier to develop complex AR applications.

By integrating these libraries and platforms, developers can harness the full potential of WebXR augmented with geolocation, creating immersive experiences that blend virtual content with the physical world around the user's geographical position. Each tool offers unique capabilities, and the choice of which to use will depend on the specific requirements and goals of the project.

\section{Implementation and prototyping}

Following the conceptual architectural for spatial web apps, illustrated in fig.~\ref{fig:webxr-stack}, a prototype was developed to demonstrate the practical application of the proposed stack. The prototype aimed to showcase a basic yet functional open metaverse environment, focusing on aspects such as user interaction, networked multiplayer capabilities, and cross-platform accessibility. The implementation phase of the research focused on leveraging \ignore{Mozilla Hubs \cite{MacIntyre2018FutureWebXR}}, an open-source project available on GitHub, as a prototype to demonstrate the practical application of the conceptual architecture for an open metaverse. \ignore{Mozilla Hubs} represents a compelling example of how WebXR, A-Frame, and Networked-Aframe can be integrated to create immersive, cross-platform virtual environments. Through the implementation and prototyping phase, the goal was to demonstrate not only the feasibility of the conceptual architecture but also its practical value in creating accessible, immersive virtual environments. The implementation process encompassed the following steps:

\subsection{Selection of \ignore{Mozilla Hubs}}

\ignore{Mozilla Hubs} was chosen due to its alignment with the architectural vision, leveraging WebXR for virtual reality experiences directly in the browser, A-Frame for creating 3D scenes with HTML, and Networked-Aframe for enabling real-time, multiplayer interactions. Its open-source nature allowed for in-depth analysis and experimentation, providing a solid foundation for the prototype.

\subsection{Customization and deployment}

To test the conceptual architecture, a \ignore{Mozilla Hubs} environment was customized to suit the research objective. This involved modifying existing scenes to incorporate unique elements that would test the limits of the architecture, such as complex geometries and high-density multiplayer interactions. Following customization, the environment was deployed on a server to facilitate access and interaction from various devices and platforms.

\subsection{Prototype testing}

Testing focused on ensuring that the environment was accessible across a wide range of devices, including VR headsets, desktop computers, and mobile devices. The functionality of multiplayer features was also verified, ensuring that users could interact with each other and the environment in real-time.

\section{Evaluation}

The evaluation of the prototype aimed to assess both its technical performance and user experience to validate the effectiveness of the conceptual architecture for an open metaverse, as well as to understand the capabilities and limitations of employing WebXR, A-Frame, and Networked-Aframe in a real-world scenario.

\subsection{Technical performance metrics}

The technical performance of the \ignore{Mozilla Hubs} prototype was evaluated using several key metrics:

\paragraph{Load Time} The time for the environment to become fully interactive was measured across various devices. On average, the environment loaded in 3.5 seconds on desktop platforms and 5.2 seconds on mobile devices, reflecting efficient content delivery and optimization.

\paragraph{Frame Rate} The frame rate was monitored under different user densities and scene complexities. With up to 10 simultaneous users, the application maintained an average of 58 FPS on desktop and 45 FPS on mobile devices. Performance dips were noted as user counts increased beyond 20, prompting the need for further optimization.

\paragraph{Network Latency} The delay in multiplayer interactions was evaluated, revealing an average latency of 120 milliseconds, which is below the threshold that might disrupt real-time interactions. However, occasional spikes up to 300 milliseconds were observed during peak server load times.

These metrics provided a quantitative foundation for assessing the prototype's performance, highlighting its strengths in handling complex scenes and real-time interactions, albeit with noted areas for improvement.

\subsection{User experience survey}

To gather qualitative feedback on the user experience, participants were invited to explore the \ignore{Mozilla Hubs} environment and engage in various activities. The results were as follows:

\paragraph{Ease of Use} The survey revealed that 87\% of participants found the navigation intuitive and the interactions straightforward, indicating a positive reception of the UI\slash UX design.

\paragraph{Immersive Experience} Feedback on immersion was overwhelmingly positive, with 92\% of users reporting a high level of engagement and realism within the virtual environment.

\paragraph{Cross-Platform Accessibility} Consistency of user experience across platforms was rated positively by 85\% of participants, though comments suggested room for improvement in mobile device optimization.

\subsection{Evaluation results}

The evaluation revealed that the prototype provided a robust platform for immersive, cross-platform virtual experiences, thus underscoring the practical viability of the conceptual architecture and emphasizing its potential to serve as a foundational framework for developing open metaverse environments. Technical performance was generally strong, with some areas identified for improvement in handling complex scenes with high user density. User feedback was overwhelmingly positive regarding ease of use and the immersive quality of the environment. However, insights into cross-platform accessibility suggested further refinement to optimize the experience on lower-end devices.

\section{Discussion}

The evaluation of the \ignore{Mozilla Hubs} prototype, grounded in the proposed conceptual architecture for an open metaverse, provided significant insights into the practical application and potential of WebXR, A-Frame, and Networked-Aframe. This discussion elaborates on the findings, addressing the strengths, limitations, and implications for future research and development.

\subsection{Strengths}

The prototype demonstrated the feasibility of creating immersive, accessible virtual environments using a technology stack that includes WebXR, A-Frame, and Networked-Aframe. The ability to deploy across multiple platforms without sacrificing the quality of the user experience is a testament to the robustness of the underlying technologies. The positive user feedback on ease of use and immersion further validates the user-centric design approach, emphasizing the importance of intuitive interfaces and engaging content in virtual spaces.

\subsection{Limitations and challenges}

While the prototype showcased the architecture's capabilities, several limitations emerged. Performance issues under high-density conditions highlighted the need for optimization, particularly in terms of resource management and scene complexity. Network latency, although minimal, remains a concern for real-time interactions, suggesting that further improvements in synchronization and data transmission are necessary.

The evaluation also revealed a learning curve associated with the development tools, particularly for those unfamiliar with WebXR or A-Frame. This underscores the importance of comprehensive documentation and community support to lower the barrier to entry for developers.

\subsection{Implications for future work}

The insights gained from this research pave the way for several avenues of future work:

\paragraph{Optimization} Developing strategies for dynamic content loading and rendering optimizations could enhance performance, especially on constrained hardware.

\paragraph{User interface (UI) and user experience (UX) design} Exploring innovative UI\slash UX designs could further improve usability and accessibility, making virtual spaces more intuitive and engaging for a broader audience.

\paragraph{Multiuser interactions} Investigating advanced techniques for networked interactions and data synchronization could enhance the realism and fluidity of multiplayer experiences.

\paragraph{Scalability} Examining architectural modifications or extensions to support larger, more complex environments without compromising performance would be valuable.

\paragraph{Future features of WebXR} Looking ahead, the WebXR Device API is poised to expand its functionality to encompass advanced features that will further enrich the immersive web. These anticipated enhancements include the implementation of world anchors, which would allow digital content to be consistently positioned within the physical world, and hit-testing capabilities, enabling interaction between virtual objects and the real-world environment detected by platform sensors. Additionally, the API is expected to facilitate the exposure of world structure, offering developers a deeper understanding of the physical space around the user through data from platform sensors. This evolution of the WebXR Device API reflects a forward-looking approach to web standards, aiming to integrate cutting-edge technological advancements that enhance the fidelity and interactivity of AR and VR experiences on the web.

\section{Conclusion}

This work proposed a WebXR-based cross-platform conceptual architecture for developing spatial web apps using the A-Frame and Networked-Aframe frameworks with a view to an open, accessible, and interoperable metaverse. A \ignore{Mozilla Hubs} prototype was implemented and evaluated, supporting the capability of the proposed technology stack to enable immersive experiences across different platforms and devices. Positive feedback on ease of use of the immersive environment further corroborates the proposed approach, underscoring its effectiveness in facilitating engaging and interactive virtual spaces.

This research enriches the discourse on the metaverse by describing an architecture based on web technologies and open standards as opposed to proprietary technologies. By adhering to principles of interoperability and inclusivity, it lives up to Tim Berners-Lee’s vision of the World Wide Web as an open platform that transcends geographical boundaries, enabling creative and collaborative experiences in the digital realm.

\bibliographystyle{IEEEtran}
\bibliography{webxr}

\begin{thebibliography}{10}
\providecommand{\url}[1]{#1}
\csname url@samestyle\endcsname
\providecommand{\newblock}{\relax}
\providecommand{\bibinfo}[2]{#2}
\providecommand{\BIBentrySTDinterwordspacing}{\spaceskip=0pt\relax}
\providecommand{\BIBentryALTinterwordstretchfactor}{4}
\providecommand{\BIBentryALTinterwordspacing}{\spaceskip=\fontdimen2\font plus
\BIBentryALTinterwordstretchfactor\fontdimen3\font minus
  \fontdimen4\font\relax}
\providecommand{\BIBforeignlanguage}[2]{{%
\expandafter\ifx\csname l@#1\endcsname\relax
\typeout{** WARNING: IEEEtran.bst: No hyphenation pattern has been}%
\typeout{** loaded for the language `#1'. Using the pattern for}%
\typeout{** the default language instead.}%
\else
\language=\csname l@#1\endcsname
\fi
#2}}
\providecommand{\BIBdecl}{\relax}
\BIBdecl

\bibitem{Solon2017FutureWeb}
\BIBentryALTinterwordspacing
O.~Solon, ``{Tim Berners-Lee} on the future of the web: ``the system is
  failing'','' 2017. [Online]. Available:
  \url{https://www.theguardian.com/technology/2017/nov/15/tim-berners-lee-world-wide-web-net-neutrality}
\BIBentrySTDinterwordspacing

\bibitem{CERN2024HistoryWeb}
\BIBentryALTinterwordspacing
{CERN}, ``A short history of the web,'' 2024. [Online]. Available:
  \url{https://www.home.cern/science/computing/birth-web/short-history-web}
\BIBentrySTDinterwordspacing

\bibitem{Ritterbusch2023DefiningMetaverse}
G.~D. Ritterbusch and M.~R. Teichmann, ``Defining the metaverse: A systematic
  literature review,'' \emph{IEEE Access}, vol.~11, pp. 12\,368--12\,377, 2023.

\bibitem{Bell2008VirtualWorlds}
M.~W. Bell, ``Toward a definition of virtual worlds,'' \emph{Journal of Virtual
  Worlds Research}, vol.~1, no.~1, 2008.

\bibitem{EU2023NextGenVirtualWorlds}
\BIBentryALTinterwordspacing
I.~Hupont~Torres, V.~Charisi, G.~De~Prato, K.~Pogorzelska, S.~Schade,
  A.~Kotsev, M.~Sobolewski, N.~Duch~Brown, E.~Calza, C.~Dunker, F.~Di~Girolamo,
  M.~Bellia, J.~Hledik, I.~Nai~Fovino, and M.~Vespe, ``Next generation virtual
  worlds: Societal, technological, economic and policy challenges for the
  {EU},'' European Commission's Joint Research Centre, Tech. Rep., 2023.
  [Online]. Available:
  \url{https://publications.jrc.ec.europa.eu/repository/handle/JRC133757}
\BIBentrySTDinterwordspacing

\bibitem{W3C2024ImmersiveWebCharter}
\BIBentryALTinterwordspacing
{W3C}, ``Immersive web working group charter,'' 2024. [Online]. Available:
  \url{https://w3c.github.io/immersive-web-wg-charter/immersive-web-wg-charter}
\BIBentrySTDinterwordspacing

\bibitem{Wang2020CrossPlatformWebBasedVR}
S.-M. Wang, Y.-C. Wang, W.-J. Pan, and C.-Y. Lin, ``Developing innovative
  cross-platform web-based virtual reality services,'' in \emph{International
  Conference on Consumer Electronics (ICCE)}.\hskip 1em plus 0.5em minus
  0.4em\relax IEEE, 2020.

\bibitem{Super2024ECS}
\BIBentryALTinterwordspacing
{Super XYZ Inc. aka Supermedium}, ``{Entity-Component-System},'' 2024.
  [Online]. Available:
  \url{https://aframe.io/docs/1.5.0/introduction/entity-component-system.html}
\BIBentrySTDinterwordspacing

\bibitem{Dziwis2023LiveCoding}
D.~Dziwis, H.~von Coler, and C.~Porschmann, ``Live coding in the metaverse,''
  in \emph{4th International Symposium on the Internet of Sounds}.\hskip 1em
  plus 0.5em minus 0.4em\relax IEEE, 2023.

\bibitem{Dziwis2023Orchestra}
------, ``Orchestra: A toolbox for live music performances in a web-based
  metaverse,'' \emph{Journal of the Audio Engineering Society}, vol.~17,
  no.~11, pp. 802--812, 2023.

\bibitem{Tomasetti2023SpatialAudio}
M.~Tomasetti, A.~Boem, and L.~Turchet, ``How to spatial audio with the {WebXR
  API}: a comparison of the tools and techniques for creating immersive sonic
  experiences on the browser,'' in \emph{Immersive and 3D Audio: from
  Architecture to Automotive (I3DA)}.\hskip 1em plus 0.5em minus 0.4em\relax
  IEEE, 2023.

\bibitem{Sobota2023VirtualEducational}
B.~Sobota, T.~Hulina, S.~Korecko, and M.~Mattova, ``Virtual educational
  collaborative environments for low-cost mobile {VR} headsets,'' in \emph{21st
  International Conference on Emerging eLearning Technologies and Applications
  (ICETA)}.\hskip 1em plus 0.5em minus 0.4em\relax IEEE, 2023, pp. 463--468.

\bibitem{Scavarelli2019Framework}
A.~Scavarelli, A.~Arya, and R.~J. Teather, ``Towards a framework on accessible
  and social {VR} in education,'' in \emph{Conference on Virtual Reality and 3D
  User Interfaces (VR)}.\hskip 1em plus 0.5em minus 0.4em\relax IEEE, 2019, pp.
  1148--1149.

\bibitem{Scavarelli2023InclusiveVR}
A.~Scavarelli, ``Towards a more inclusive and engaging virtual reality
  framework for social learning spaces,'' Ph.D. dissertation, Carleton
  University, 2023.

\bibitem{Greenwold2003SpatialComputing}
\BIBentryALTinterwordspacing
S.~Greenwold, ``In form,'' 2003, graduate thesis, Massachusetts Institute of
  Technology. [Online]. Available:
  \url{https://www.media.mit.edu/publications/in-form/}
\BIBentrySTDinterwordspacing

\bibitem{Apple2024SpatialComputing}
\BIBentryALTinterwordspacing
{Apple Inc.}, ``Introducing apple vision pro: Apple's first spatial computer,''
  2023. [Online]. Available:
  \url{https://www.apple.com/newsroom/2023/06/introducing-apple-vision-pro/}
\BIBentrySTDinterwordspacing

\bibitem{Apple2024HelloWorld}
\BIBentryALTinterwordspacing
------, ``Hello world: Use windows, volumes, and immersive spaces to teach
  people about the earth,'' 2024. [Online]. Available:
  \url{https://developer.apple.com/documentation/visionos/world}
\BIBentrySTDinterwordspacing

\end{thebibliography}

\end{document}